\pdfoutput=1

\documentclass[11pt]{article}

\usepackage[]{emnlp2021}

\usepackage{times}
\usepackage{latexsym}

\usepackage[T1]{fontenc}

\usepackage[utf8]{inputenc}

\usepackage{microtype}

\usepackage{paralist}
\usepackage{graphicx}
\usepackage{makecell}
\usepackage{multirow}
\usepackage{hyperref}
\usepackage{amsmath}
\usepackage{xspace}

\newcommand{\model}{Graformer\xspace}
\title{Multilingual Translation via Grafting Pre-trained Language Models}

\author{Zewei Sun\textsuperscript{\rm 1}, Mingxuan Wang\textsuperscript{\rm 1}, Lei Li\textsuperscript{\rm 2}\thanks{ ~~Work is done while at ByteDance.} \\
  \textsuperscript{\rm 1} ByteDance AI Lab\\
  \texttt{\{sunzewei.v,wangmingxuan.89\}@bytedance.com}, \\
  \textsuperscript{\rm 2} University of California, Santa Barbara\\
   \texttt{lilei@cs.ucsb.edu}} 

\begin{document}
\maketitle

\begin{abstract}

Can pre-trained BERT for one language and GPT for another be glued together to translate texts? Self-supervised training using only monolingual data has led to the success of pre-trained (masked) language models in many NLP tasks. However, directly connecting BERT as an encoder and GPT as a decoder can be challenging in machine translation, for GPT-like models lack a cross-attention component that is needed in seq2seq decoders. In this paper, we propose \textbf{\model} to graft separately pre-trained (masked) language models for machine translation. 
With monolingual data for pre-training and parallel data for grafting training, we maximally take advantage of the usage of both types of data. Experiments on 60 directions show that our method achieves average improvements of 5.8 BLEU in x2en and 2.9 BLEU in en2x directions comparing with the multilingual Transformer of the same size\footnote{Our code will be public in \url{https://github.com/sunzewei2715/Graformer}}.

\end{abstract}

\section{Introduction}
\label{sec:intro}


In recent years, pre-trained (masked) language models have achieved significant progress in all kinds of NLP tasks~\cite{devlin2019bert,radford2019language}. Among them, neural machine translation (NMT) is also explored by several attempts~\cite{yang2020towards,zhu2020incorporating,rothe2020leveraging}. The pre-training and fine-tuning style becomes an important alternative to take advantage of monolingual data~\cite{yang2020csp,yang2020alternating,liu2020multilingual,pan2021contrastive}.

\begin{figure}[t]
    \centering
    \includegraphics[width=0.5\textwidth]{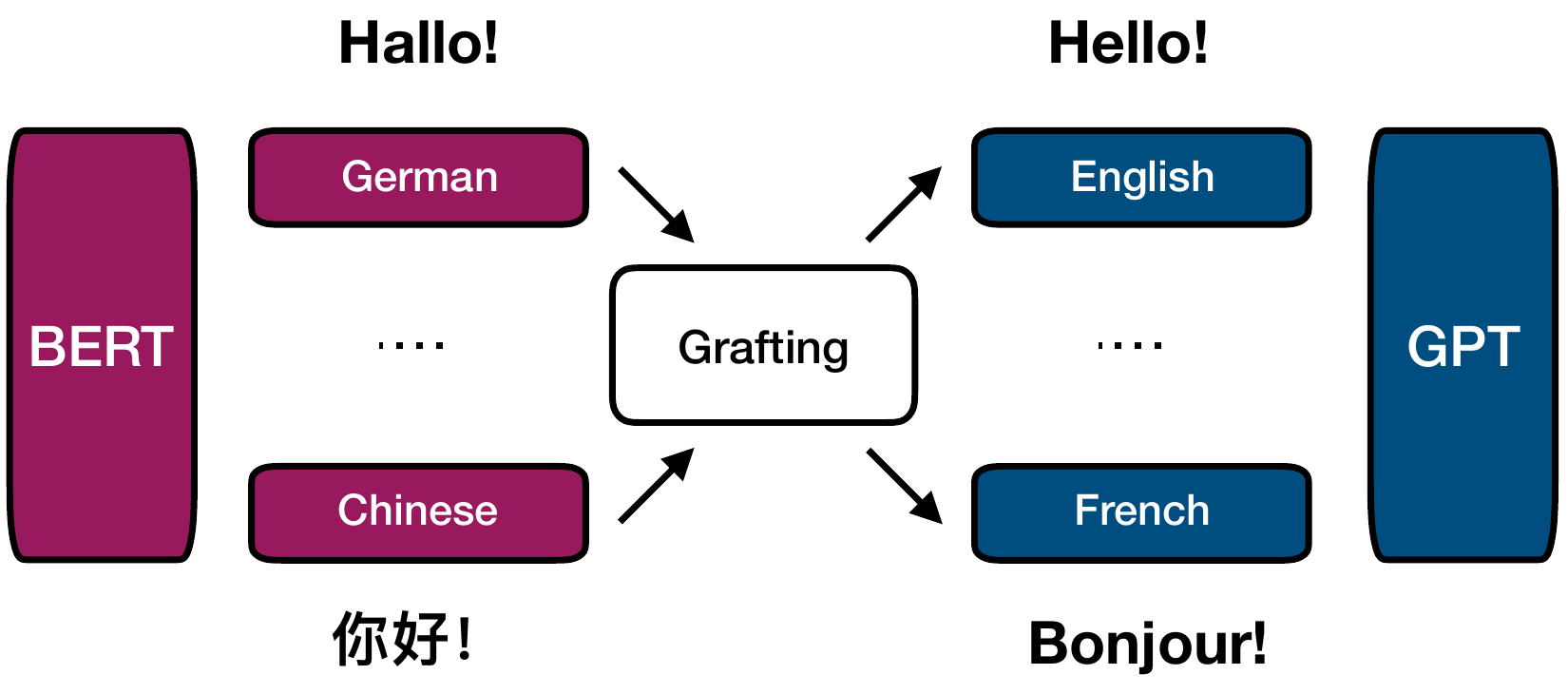}
    \caption{Grafting pre-trained (masked) language models like BERT and GPT for machine translation.}
    \label{fig:intuition}
\end{figure}

An intuitive question comes as: Can we bridge BERT-like pre-trained encoders and GPT-like decoders to form a high-quality translation model? Since they only need monolingual data, we can reduce the reliance on the large parallel corpus. 
Moreover, if the combination of models is universal, it can be applied to translation for multiple languages, as is shown in Figure~\ref{fig:intuition}.

However, though many works successfully gain improvements by loading encoder/decoder parameters from BERT-like pre-trained encoders~\cite{zhu2020incorporating,guo2020incorporating}, they do not achieve satisfactory results with loading decoder parameters from GPT-like pre-trained decoders~\cite{yang2020towards,rothe2020leveraging}. Theoretically, the well-trained decoder model like GPT should bring better generation ability to the translation model. We suggest the outcome may be attributed to the architecture mismatch.

\begin{figure}[h]
    \centering
    \includegraphics[width=0.4\textwidth]{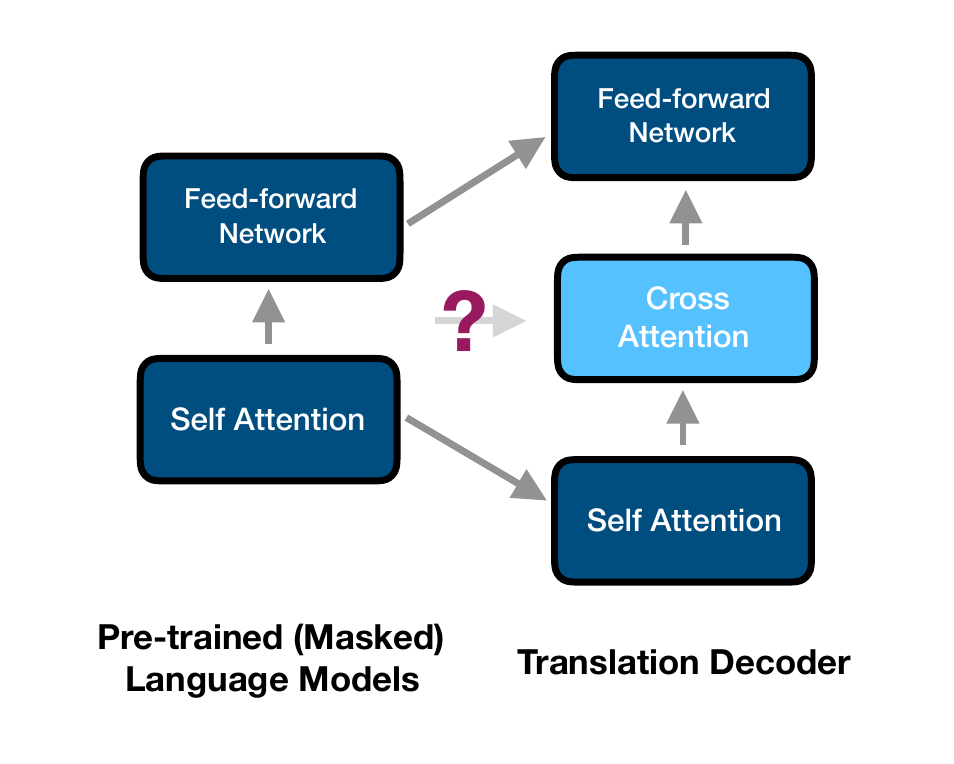}
    \caption{Taking the popular architecture Transformer~\cite{vaswani2017attention} as an example, the translation model has a ``cross-attention'' sub-layer, while pre-trained (masked) language models have none.}
    \captionsetup{font={scriptsize}}
    \label{fig:intro}
\end{figure}

Pre-trained (masked) language models predict the current word solely based on the internal context while the translation decoder has to capture the source context. Specifically, the decoder in NMT has a ``cross-attention'' sub-layer that plays a transduction role~\cite{bahdanau2015neural}, while pre-trained models have none, as is shown in Figure~\ref{fig:intro}. This mismatch between the generation models and conditional generation models makes it a challenge for the usage of pre-trained models as translation decoders.

Therefore, some previous works manually insert cross-attention sub-layer or adapters~\cite{rothe2020leveraging,ma2020xlm,guo2020incorporating}. However, the extra implantation may influence the ability of the pre-trained model. Other works try to avoid this problem by directly pre-training a seq2seq model and conduct fine-tuning~\cite{tang2020multilingual,yang2020alternating,luo2020veco}. However, the pre-training objective is usually a variant of auto-encoding~\cite{song2019mass,liu2020multilingual}, which is different from the downstream translation objective and may not achieve adequate improvements~\cite{lin2020pre}. 

In this paper, we mainly focus on exploring the best way to simultaneously take advantage of the pre-trained representation model and generation model (e.g., BERT+GPT) without limiting their strengths. The primary target is to link the generation model to the source side and maintain the invariability of the architecture in the meantime. Therefore, we propose \textbf{\model}, with pre-trained models grafted by a connective sub-module. The structure of the pre-trained parts remains unchanged, and we train the grafting part to learn to translate. For universality and generalization, we also extend the model to multilingual NMT, achieving mBERT+mGPT.

Generally, the translation process can be divided into three parts: representation, transduction, and generation, respectively achieved by the encoder, cross-attention, and decoder. In multilingual NMT, the transduction can only be trained with multiple parallel data. But the rest two can be pre-trained with multiple monolingual data, which is tens or hundreds of the size of parallel one. To maximize the efficacy of each part, we firstly pre-train a multilingual BERT and multilingual GPT. Then they are grafted to implement translation. With the architecture consistency, we can reserve the language knowledge of the pre-trained models and obtain a strong translation model flexibly at the same time.

Experiments on 30 language directions show that our method improves the results of multilingual NMT by 2.9 and 5.8 BLEU on average. It also achieves gains of 9.2 to 13.4 BLEU scores on zero-shot translation settings. 
In addition, it verifies that such translation capability can be well transferred to other languages without fine-tuning on the target parallel corpus.

\section{Related Work}
\label{sec:related}

This paper is related to a chain of studies of multilingual translation and pre-trained models. 

\subsection{Multilingual Neural Machine Translation}

With the development of NMT, multilingual neural machine translation (MNMT) also attracts a great amount of attention. 
\newcite{dong2015multi,firat2016multi,firat2016zero} take early attempts and confirm its feasibility. 
The most well-known work is from~\newcite{johnson2017google}, who conduct a series of interesting experiments. And the usage of the language token style is widely accepted. 
Also, many subsequent works continuously explore new approaches in MNMT, such as parameter sharing~\cite{blackwood2018multilingual,wang2019compact,tan2019multilingual}, parameter generation~\cite{platanios2018contextual}, knowledge distillation~\cite{tan2019multilingualkd}, learning better representation~\cite{wang2019multilingual}, massive training~\cite{aharoni2019massively,arivazhagan2019massively}, interlingua~\cite{zhu2020language}, and adpater~\cite{zhu2021serial}.
These works mainly utilize parallel data.

There are also some works taking advantage of monolingual corpus.
\newcite{zhang2020improving,wang2020multi} use back-translation (BT) to improve MNMT. However, for MNMT, BT is tremendously costly, reaching $O(n)$, or even $O(n^2)$.
\newcite{siddhant2020leveraging,wang2020multi} adopt multi-task learning (MTL), combining with other tasks such as masked language model (MLM)~\cite{devlin2019bert}, denoising auto-encoding (DAE)~\cite{vincent2008extracting}, or masked sequence-to-sequence generation (MASS)~\cite{song2019mass}. However, the optimization target is different from translation, which may interfere with the training and limit the usage of extremely large-scale monolingual data.

\subsection{Pre-trained Models}

In recent years, pre-train models have become very popular in both research and industry communities. With downstream fine-tuning, plenty of significant results are achieved in NLP field~\cite{qiu2020pre}. 

\newcite{devlin2019bert,liu2019roberta,conneau2019cross,conneau2020unsupervised} take masked language model (MLM) as the training target. The input tokens are randomly masked, and the model learns the representation by maximizing their likelihood.
\newcite{radford2018improving,radford2019language,brown2020language} use language model (LM) as their learning goal. With historical contexts, the model acquires language knowledge by learning to predict the next word.
\newcite{raffel2020exploring,xue2020mt5,lewis2020bart,liu2020multilingual,lin2020pre} choose direct sequence-to-sequence (seq2seq) for training. The pre-train tasks can be machine translation, question answering, classification, etc.

\subsection{Pre-trained Models for NMT}
\label{subsec:related}

Since pre-trained models can significantly boost relevant tasks, several recent studies try to combine them with NMT. They can be roughly divided into two groups, depending on whether the models are pre-trained uniformly or separately. 

\subsubsection{United Style}
The first category is pre-training directly on seq2seq tasks and providing downstream MT with consistent architectures.
\newcite{tang2020multilingual} tune translation models from a pre-trained seq2seq model, mBART~\cite{liu2020multilingual}, and obtain significant improvements.
\newcite{yang2020csp} pre-train a seq2seq model with some input tokens replaced by another language from lexicon induction.
\newcite{luo2020veco} pre-train the encoder and decoder in a single model that shares parameters. Then the parameters are partially extracted for tuning, depending on the tasks (NLU or NLG).

However, the pre-training objective of these works is usually a variant of auto-encoding~\cite{song2019mass,liu2020multilingual}, which is different from the downstream translation objective and may not achieve adequate improvements~\cite{lin2020pre}. 

\subsubsection{Fused Style}
The second category is pre-training the encoder or decoder independently and fusing them with the translation model in the fine-tuning stage.
\newcite{yang2020towards,zhu2020incorporating,guo2020incorporating,ma2020xlm} fuse BERT/RoBERTa into NMT with extra encoders or adapters. \newcite{yang2020alternating} propose alternating language modeling as the target of the pre-trained encoder. \newcite{rothe2020leveraging} explore the usage of GPT but still manually insert extra cross-attention. \newcite{weng2020acquiring} use dynamic fusion mechanism and knowledge distillation to integrate the representation of the pre-trained models into NMT models. 

These works either do not touch the decoder side or modify the architecture and conduct fine-tuning to fuse BERT/GPT into the decoder model. As mentioned in Section~\ref{sec:intro}, the modification of the model architecture may influence the model ability and harm the performance.

\section{Approach}
\label{sec:approach}

\begin{figure}
    \centering
    \includegraphics[width=0.5\textwidth]{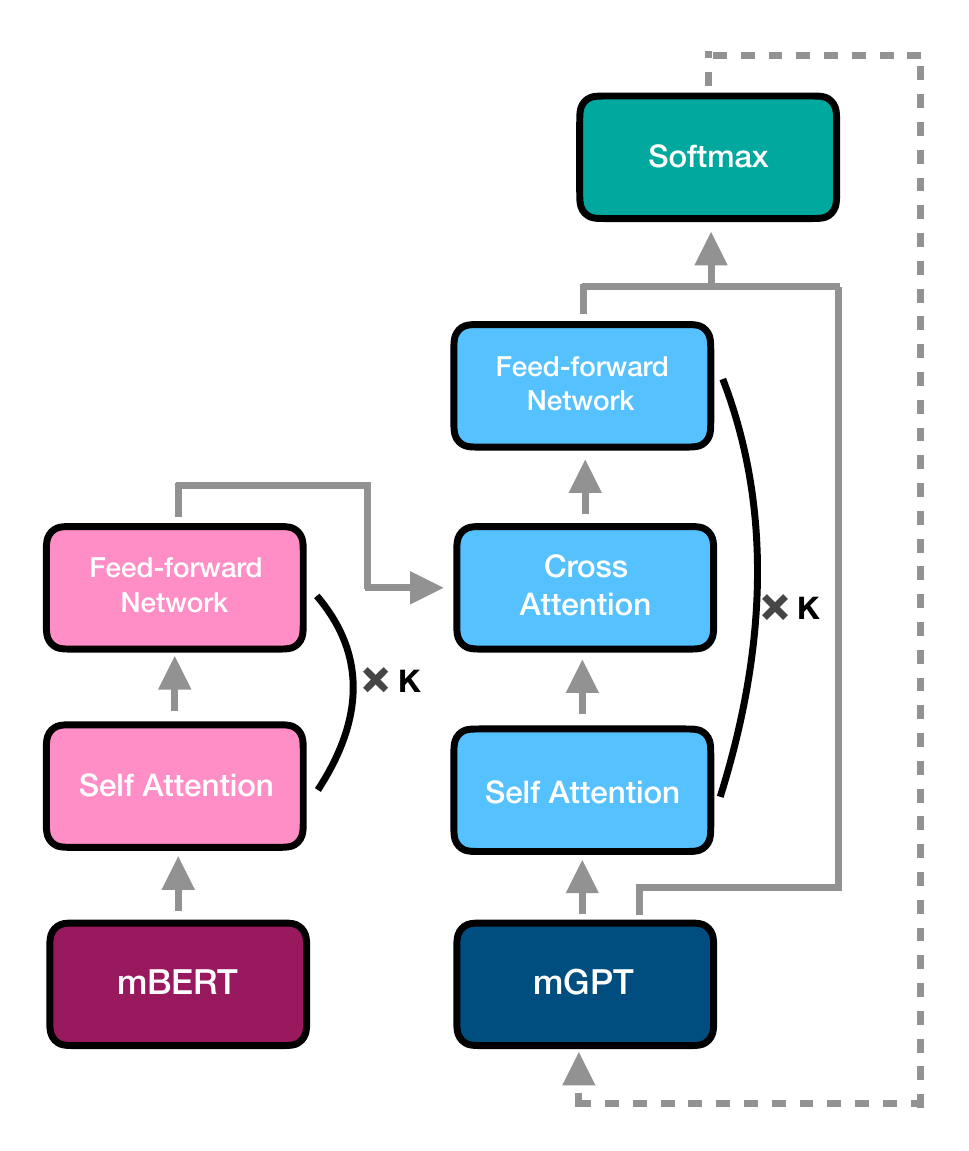}
    \caption{The model architecture of \model. The pre-trained multilingual encoder (mBERT) and decoder (mGPT) are grafted to achieve multilingual translation. The dashed line means feeding in the last token.}
    \label{fig:bridge}
\end{figure}

To maintain the original model structure of pre-trained models, we propose \model, as is in Figure~\ref{fig:bridge}.
For the encoder side, we stack another $K$-layers encoder ($K=6$, in this paper) on pre-trained mBERT to help it adapt to the translation training. For the decoder side, we do similarly, except we append cross-attention layers to extract conditional context from the source. 
Unlike previous works, we maintain the integrality of mBERT and mGPT and do not change their architectures.

Finally, we employ a residual connection~\cite{he2016deep} that we combine the hidden state outputted by mGPT and the grafting decoder. The summed context is then fed into the softmax layer. This integration is for utilizing the generation ability of the pre-trained decoder to help to generate a better language model. 

As mentioned in Section~\ref{sec:intro}, we try to take advantage of both multiple parallel data and multiple monolingual data so as to maximize the efficacy of representation, transduction, and generation, respectively. Therefore, our training methods can be separated into two stages: \textit{1)} pre-train on the multiple monolingual data and obtain independent encoder (representation) and decoder (generation); \textit{2)} fine-tune on the multilingual parallel data to graft two models (transduction).

\subsection{Pre-train Multilingual BERT \ \ \ \ \  (Encoder for Representation)}

Inspired by~\newcite{devlin2019bert,liu2019roberta,conneau2019cross,conneau2020unsupervised}, we use masked language model (MLM) as the training goal with the masked probability of 15\%. Specifically, we adopt Transformer~\cite{vaswani2017attention} encoder with $N$ layers ($N=6$, in this paper). To make cross-lingual token representation more universal, we add no language token as previous works do. The training goal is as follows: 

\begin{equation}
    \mathcal{L}_{MLM} = -\sum_{\hat{x}\in m(\mathbf{x})}{\text{log}\ p(\hat{x}|\mathbf{x}_{\backslash m(\mathbf{x})})}
\end{equation}

$m(\mathbf{x})$ and $\backslash m(\mathbf{x})$ denote the masked words and rest words from $\mathbf{x}$

\subsection{Pre-train Multilingual GPT \ \ \ \ \ \ \ \ \  (Decoder for Generation)}

Inspired by~\newcite{radford2018improving,radford2019language,brown2020language}, we use auto-regressive language model (LM) as the training goal. Specifically, we adopt Transformer~\cite{vaswani2017attention} decoder with $N$ layers ($N=6$, in this paper). To specify the generation language, we set a unique language token (e.g., \texttt{<2en>}) as the first input for the language model. The training goal is as follows: 

\begin{equation}
    \mathcal{L}_{LM} = -\sum_{t=1}^{T}{\text{log}\ p(x_t|\mathbf{x}_{<t})}
\end{equation}

$T$ denotes the length of sequence. $\mathbf{x}_{<t}=\texttt{<2lang>}, x_1, x_2, ..., x_{t-1}$.

\subsection{Fine-tune Multilingual Translation (Grafting for Transduction)}

After obtaining the pre-trained encoder and decoder, we tune the model to link the representation model and generation model. The training goal is as follows: 

\begin{equation}
    \begin{aligned}
        \mathcal{L}_{MT} = \ & \text{softmax}(W_{o_{1}}h_{N} \ +\ W_{o_{2}}h_{N+K})
    \end{aligned}
\end{equation}

$h_{N}$ denotes the hidden state of the last layer in mGPT. $h_{N+K}$ denotes the hidden state of the last layer in the grafting decoder. $W_{o_{1}}$ and $W_{o_{2}}$ denote the corresponding output matrix. The former one shares the same parameters with the target-side embedding.

In the tuning stage, we freeze the pre-trained decoder parameters (including $W_{o_{1}}$) and tune the grafting parameters as well as the pre-trained encoder. Our ablation study shows that this setting yields the best performance, as is in the experiment section.

\section{Experiments}
\label{sec:experiments}

\begin{table*}[ht]
    \centering
    \begin{tabular}{l|cccccccccc}
        \Xhline{3\arrayrulewidth}
        Model & bg & bn & bs & cs & de & el & es & et & fa & fi \\ \hline
        Transformer & 32.0 & 12.5 & 30.3 & 23.7 & 28.7 & 30.7 & 34.7 & 17.5 & 20.4 & 17.1 \\
        mBART & - & - & - & 26.4 & 32.8 & - & 38.1 & 20.9 & - & 19.9 \\
        \model & \textbf{38.5} & \textbf{18.1} & \textbf{36.5} & \textbf{29.4} & \textbf{35.5} & \textbf{37.4} & \textbf{40.7} & \textbf{24.0} & \textbf{26.9} & \textbf{23.0} \\
        \Xhline{3\arrayrulewidth}
        Model & fr & hi & hr & hu & it & ja & kk & lt & mk & mr \\ \hline
        Transformer & 33.1 & 18.7 & 30.4 & 19.8 & 31.3 & 10.1 & 7.6 & 20.1 & 29.8 & 9.4 \\
        mBART & 36.5 & 22.9 & - & - & 34.7 & 12.0 & 8.9 & 23.6 & - & - \\
        \model & \textbf{39.2} & \textbf{25.1} & \textbf{36.7} & \textbf{26.1} & \textbf{37.2} & \textbf{13.7} & \textbf{10.5} & \textbf{27.2} & \textbf{35.7} & \textbf{13.0} \\
        \Xhline{3\arrayrulewidth}
        Model & nl & pl & pt & ro & ru & sr & ta & tr & uk & zh \\ \hline
        Transformer & 28.9 & 19.7 & 34.8 & 28.6 & 20.8 & 29.0 & 5.8 & 18.7 & 23.4 & 15.6 \\
        mBART & 32.9 & - & - & 32.2 & 22.6 & - & - & 22.6 & - & 18.1 \\
        \model & \textbf{35.2} & \textbf{25.1} & \textbf{41.5} & \textbf{35.1} & \textbf{25.1} & \textbf{35.6} & \textbf{10.2} & \textbf{25.5} & \textbf{28.9} & \textbf{19.9} \\
        \Xhline{3\arrayrulewidth}
    \end{tabular}
    \caption{The results of x$\rightarrow$en directions, with average improvements of 5.8 against baseline (22.8$\rightarrow$28.6)}
    \label{tab:x2en}
\end{table*}

\begin{table*}[ht]
    \centering
    \begin{tabular}{l|cccccccccc}
        \Xhline{3\arrayrulewidth}
        Model & bg & bn & bs & cs & de & el & es & et & fa & fi \\ \hline
        Transformer & 28.8 & 11.3 & 23.4 & 16.6 & 23.7 & 25.9 & 33.0 & 14.0 & 12.5 & 12.1 \\
        mBART & - & - & - & 17.7 & 25.8 & - & 35.2 & 14.1 & - & 13.2 \\
        \model & \textbf{33.0} & \textbf{14.1} & \textbf{26.3} & \textbf{20.2} & \textbf{27.8} & \textbf{29.8} & \textbf{37.5} & \textbf{16.1} & \textbf{14.2} & \textbf{14.4} \\
        \Xhline{3\arrayrulewidth}
        Model & fr & hi & hr & hu & it & ja & kk & lt & mk & mr \\ \hline
        Transformer & 33.5 & 15.3 & 23.2 & 14.7 & 28.9 & 11.1 & 3.4 & 12.8 & 22.2 & 9.3 \\
        mBART & 35.8 & 16.5 & - & - & 30.6 & 12.6 & 3.0 & 14.2 & - & - \\
        \model & \textbf{37.8} & \textbf{18.1} & \textbf{26.8} & \textbf{17.2} & \textbf{32.5} & \textbf{12.8} & \textbf{3.8} & \textbf{15.9} & \textbf{25.7} & \textbf{10.6} \\
        \Xhline{3\arrayrulewidth}
        Model & nl & pl & pt & ro & ru & sr & ta & tr & uk & zh \\ \hline
        Transformer & 25.9 & 12.8 & 32.0 & 24.7 & 16.1 & 18.7 & 13.6 & 11.6 & 17.3 & 21.2 \\
        mBART & 28.9 & - & - & 27.1 & 16.9 & - & - & \textbf{13.4} & - & 22.2 \\
        \model & \textbf{29.0} & \textbf{15.8} & \textbf{36.6} & \textbf{29.1} & \textbf{19.0} & \textbf{21.4} & \textbf{14.7} & 13.3 & \textbf{19.5} & \textbf{23.0} \\
        \Xhline{3\arrayrulewidth}
    \end{tabular}
    \caption{The results of en$\rightarrow$x directions, with average improvements of 2.9 against baseline (19.0$\rightarrow$21.9)}
    \label{tab:en2x}
\end{table*}

In this paper, we perform many-to-many style multilingual translation~\cite{johnson2017google}. The detailed illustrations of the datasets and implementation are as follows.

\subsection{Datasets and Preprocess}
\begin{compactitem}
    \item \textbf{Pre-training:} We use News-Crawl corpus~\footnote{\url{http://data.statmt.org/news-crawl}} plus WMT datasets. We conduct deduplication and label the data by language. In the end, we collect 1.4 billion sentences in 45 languages, which is only one-fifth of that of mBART~\cite{liu2020multilingual}. The detailed list of languages and corresponding scales is in Appendix~\ref{apdx:langs}.
    \item \textbf{Multilingual Translation:} We use TED datasets, the most widely used MNMT datasets, following~\newcite{qi2018and,aharoni2019massively}. We extract 30 languages~\footnote{We use the corpus of ``zh\_cn'' instead of ``zh''.} from \& to English, with the size of 3.18M sentence pairs in raw data and 10.1M sentence pairs in sampled bidirectional data. The detailed list of language pairs and scales is in Appendix~\ref{apdx:langs}.
    We download the data from the open source~\footnote{\url{https://github.com/neulab/word-embeddings-for-nmt}} and conduct detokenization with Moses Detokenizer~\cite{koehn2007moses}~\footnote{\url{https://github.com/moses-smt/mosesdecoder/blob/master/scripts/tokenizer/detokenizer.perl}}.
    \item \textbf{Zero-shot and Bilingual Translation:} We use WMT 2014 German-English (4.5M sentence pairs) and French-English (36M sentence pairs) datasets.
    \item \textbf{Sample:} Upsampling is an important way to improve the performance of low-resource pairs~\cite{arivazhagan2019massively}. Therefore, sentences are sampled according to a multinomial distribution with probabilities $\{q_i\}$, where $q_i \propto p_i^{\alpha}$, $p_i$ is the proportion of language$_i$.
    For monolingual pre-training, we follow~\cite{conneau2019cross,liu2020multilingual} and set $\alpha=0.7$. For parallel fine-tuning, we follow~\cite{arivazhagan2019massively} and and set $\alpha=0.2$ ($T=5$).
    \item \textbf{Tokenization:} Like previous works, we use sentencepiece~\cite{kudo2018sentencepiece} and learn a joint vocabulary of 64000 tokens.
\end{compactitem}

\subsection{Implementation Details}

\begin{compactitem}
    \item \textbf{Architecture:} We use Transformer~\cite{vaswani2017attention} as our basic structure with pre-norm style~\cite{xiong2020layer}, and GELU~\cite{hendrycks2016gaussian} as activation function. Specifically, we adopt 1024 dimensions for the hidden state, 4096 dimensions for the middle FFN layer, and 16 heads for multi-head attention. Learnable position embedding is also employed. For baseline models, we use 12 layers. For pre-trained ones, we use Transformer encoder and decoder (without cross-attention) with 6 layers, respectively. For the grafting part, we add another 6 layers.
    \item \textbf{Training:} We train the models with a batch size of 320,000 tokens on 16 Tesla V100 GPUs. For pre-training, we go through the total data for five times. Parameters are optimized by using Adam optimizer~\cite{kingma2015adam}, with $\beta_1=0.9$, $\beta_2=0.98$, with $warmup\_steps=4000$. Without extra statement, we use dropout $=0.3$~\cite{srivastava2014dropout}. Label smoothing \cite{szegedy2016rethinking} of value $=0.1$ is also adopted. Besides, we use fp16 mixed precision training~\cite{micikevicius2018mixed} with Horovod library with RDMA inter-GPU communication~\cite{sergeev2018horovod}. 
    \item \textbf{Evaluation:} We uniformly conduct beam search with size $=5$ and length penalty $\alpha=0.6$. For hi, ja, and zh, we use SacreBLEU~\cite{post2018call}. Otherwise, we use tokenized BLEU~\cite{papineni2002bleu} with the open-source script~\footnote{\url{https://github.com/pytorch/fairseq/blob/master/examples/m2m_100/tok.sh}}.
\end{compactitem}

\subsection{Main Results}

As is shown in Table~\ref{tab:x2en} and \ref{tab:en2x}, our methods obtain significant improvements across all language pairs. For x$\rightarrow$en and en$\rightarrow$x pairs, advances of nearly 6 BLEU and 3 BLEU are achieved. We also compare the results with loading from mBART, a well-known multilingual pre-trained sequence-to-sequence model~\cite{liu2020multilingual}~\footnote{\url{https://dl.fbaipublicfiles.com/fairseq/models/mbart/mbart.cc25.v2.tar.gz}}. Due to the language difference, we only tune the model on a part of languages. With both 12-layers depth and 1024-dimensions width, our method outperforms mBART on almost all pairs, proving the superiority of \model comparing with pre-training in United Style mentioned in Section~\ref{sec:related}. It is worth noticing that we only use the one-fifth amount of the data of mBART. 

\subsection{Ablation Study}

To verify the contribution of each part of our model, we do a series of ablation studies. As is shown in Table~\ref{tab:ablation} and~\ref{tab:layer}, we can draw at least four empirical conclusions.

\textbf{Encoder needs tuning, decoder needs not.} In Table~\ref{tab:ablation}, comparing Row 1 with Row 2, and Row 5 with Row 8, we can see that the tuning of the encoder is essential. It can bring further improvements. However, freezing pre-trained decoder parameters is a better choice. Comparing Row 3 with Row 4, and Row 6 with Row 8, we can see that tuning may lead to a drop for decoder. It seems that the pre-trained decoder model learns much more knowledge, and its original language model can better guide the generation.

\textbf{Decoder matters more.} In Table~\ref{tab:ablation}, comparing Row 1,2,3,4, we can see that the pre-trained decoder yields more progress than the pre-trained encoder. This shows that involving only pre-trained encoders like BERT into MT is limited. The performance can be further enhanced with the introduction of pre-trained decoders.

\textbf{Residual connection contributes.} In Table~\ref{tab:ablation}, comparing Row 7 with Row 8, we can see that the residual connection from the pre-trained decoder can further boost the results. The well-trained language model effectively helps the translation model. It also shows the importance of incorporating the knowledge-rich generation model.

\begin{table}[h]
    \centering
    \begin{tabular}{c|c|c|cc}
        \Xhline{3\arrayrulewidth}
        Row & Encoder & Decoder & x$\rightarrow$en & en$\rightarrow$x \\
        \Xhline{3\arrayrulewidth}
        0 & - & - & 22.8 & 19.0 \\
        1 & Freeze	& -	& 23.2 & 19.2 \\
        2 & Fine-tune & - & 27.0 & 20.2 \\
        3 & - & Freeze & 27.8 & 21.0 \\
        4 & - & Fine-tune & 25.2 & 19.9 \\
        5 & Freeze & Freeze & 25.8 & 20.4 \\
        6 & Fine-tune & Fine-tune & 27.0 & 19.4 \\
        7 & Fine-tune & Freeze* & 28.1 & 20.9 \\
        8 & Fine-tune & Freeze & \textbf{28.6} & \textbf{21.9} \\
        \Xhline{3\arrayrulewidth}
    \end{tabular}
    \captionsetup{font={footnotesize}}
    \caption{Each number is the average BLEU of 30 language directions. ``-'' means not loading from pre-trained models. ``*'' means the residual connection is abandoned.}
    \label{tab:ablation}
\end{table}


\begin{table}[h]
    \centering
    \begin{tabular}{c|c|cc}
        \Xhline{3\arrayrulewidth}
        Encoder & Decoder & x$\rightarrow$en & en$\rightarrow$x \\
        \Xhline{3\arrayrulewidth}
        6+6 & 6+6 & 28.6 & 21.9 \\
        \Xhline{1\arrayrulewidth}
        6+6 & 6+5 & 28.7 & 21.7 \\
        6+6 & 6+4 & 28.2 & 21.6 \\
        6+6 & 6+3 & 28.3 & 21.6 \\
        6+6 & 6+2 & 28.2 & 21.1 \\
        6+6 & 6+1 & 27.9 & 18.0 \\
        \Xhline{1\arrayrulewidth}
        6+5 & 6+6 & 28.6 & 21.3 \\
        6+4 & 6+6 & 28.5 & 21.5 \\
        6+3 & 6+6 & 28.4 & 21.7 \\
        6+2 & 6+6 & 28.4 & 21.0 \\
        6+1 & 6+6 & 28.0 & 20.8 \\
        6 & 6+6 & 28.0 & 20.7 \\
        \Xhline{3\arrayrulewidth}
    \end{tabular}
    \caption{Each number is the average BLEU of 30 language directions. ``x+y'' means the combination of x-layers pre-trained (masked) language models and y-layers grafting models.}
    \label{tab:layer}
\end{table}

\textbf{Layer number has slight effects.} In Table~\ref{tab:layer}, as the number of layers decreases, the performance drops slightly for both the encoder and decoder. But the extent of the decline is limited. Even no extra encoder layer or one-layer extra decoder can maintain a relatively high performance.

\subsection{Well-trained Language Model Helps}

Except for BLEU, we also study how the pre-trained generation model influence the translation model. We speculate that the pre-trained decoder helps to translate through combining the well-trained language model. Therefore, we collect and compare the perplexity of the models on the validation sets.

As is in Table~\ref{tab:ppl}, we can see that our method significantly lowers the perplexity comparing to the baseline model. The pre-trained decoder brings in better representation and language knowledge. Also, the residual connection from the original pre-trained decoder can further improve the results, illustrating the enlightening role the well-trained language model plays.

\begin{table}[h]
    \centering
    \begin{tabular}{l|cc}
        \Xhline{3\arrayrulewidth}
        Model & x$\rightarrow$en & en$\rightarrow$x \\
        \hline
        Transformer & 8.64 & 8.76 \\
        \model* & 5.60 & 6.58 \\
        \model & \textbf{5.27} & \textbf{6.21} \\
        \Xhline{3\arrayrulewidth}
    \end{tabular}
    \caption{The perplexity of models. Each number is the average result of 30 language directions. ``*'' means the residual connection is abandoned.}
    \label{tab:ppl}
\end{table}

\subsection{Better than Fused Styles}

\begin{figure}[h]
    \centering
    \includegraphics[width=0.5\textwidth]{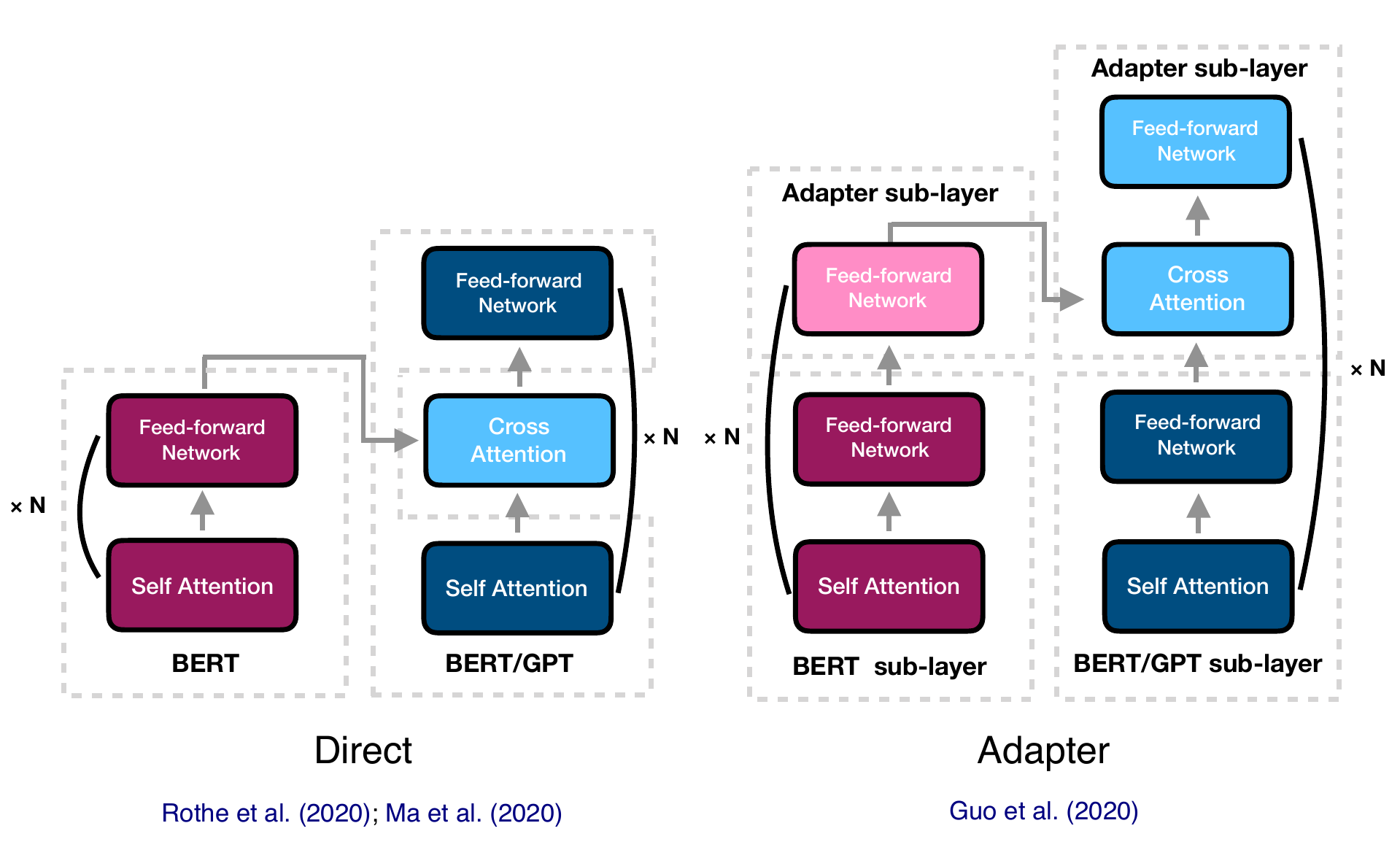}
    \caption{The model architecture of ``Direct'' (left)~\cite{rothe2020leveraging,ma2020xlm} and ``Adapter'' (right)~\cite{guo2020incorporating}.}
    \label{fig:others}
\end{figure}

Besides \textit{United Style} (mBART), we also compare our method with \textit{Fused Style}. Specifically, we choose two typical works, as are in Figure~\ref{fig:others}: \textit{1)} loading parameters directly and ignoring cross-attention (denoted as ``Direct'')~\cite{rothe2020leveraging,ma2020xlm}; \textit{2)} insert extra cross-attention layers into each decoder sub-layer and freeze pre-trained models (denoted as ``Adapter'')~\cite{guo2020incorporating}. We re-implement the models with the same depth and width as \model.

\begin{table}[h]
    \centering
    \begin{tabular}{l|cc|cc}
        \Xhline{3\arrayrulewidth}
        \multirow{2}{*}{Model} & \multicolumn{2}{c|}{BLEU$\uparrow$} & \multicolumn{2}{c}{Perplexity$\downarrow$} \\ 
        \cline{2-5}
        & x$\rightarrow$en & en$\rightarrow$x & x$\rightarrow$en & en$\rightarrow$x \\ 
        \hline
        Direct & 27.1 & 20.5 & 6.61 & 8.06 \\ 
        Adapter & 27.4 & 19.8 & 5.78 & 6.71 \\
        \model & \textbf{28.6} & \textbf{21.9} & \textbf{5.27} & \textbf{6.21} \\
        \Xhline{3\arrayrulewidth}
    \end{tabular}
    \caption{Each number is the average BLEU/Perplexity of 30 language directions. Our model
    outperform related methods in fused style.}
    \label{tab:model}
\end{table}

The crucial difference is that we leave the pre-trained decoder module unchanged and complete. Other works inject extra layers internally, such as cross-attention or adapters.
Specifically, they go like $layer_1 \rightarrow adapter_1 \rightarrow layer_2 \rightarrow adapter_2 \rightarrow ... \rightarrow layer_{N} \rightarrow adapter_{N}$. The well-trained bond between $layer_i$ and $layer_{i+1}$ is broken, which can not activate the full potential of the pre-trained decoder.

Differently, we maintain the original structure and even feed its output into the final layer. These strategies are all for the sake of fully taking advantage of the pre-trained generation model. As is in Table~\ref{tab:model}, our approach outperforms other two methods (The detailed results are in Appendix~\ref{apdx:assembled}).

\subsection{\model Maintains Good Performance in Few-Shot Translation}

\begin{figure}[ht]
    \centering
    \includegraphics[width=0.495\textwidth]{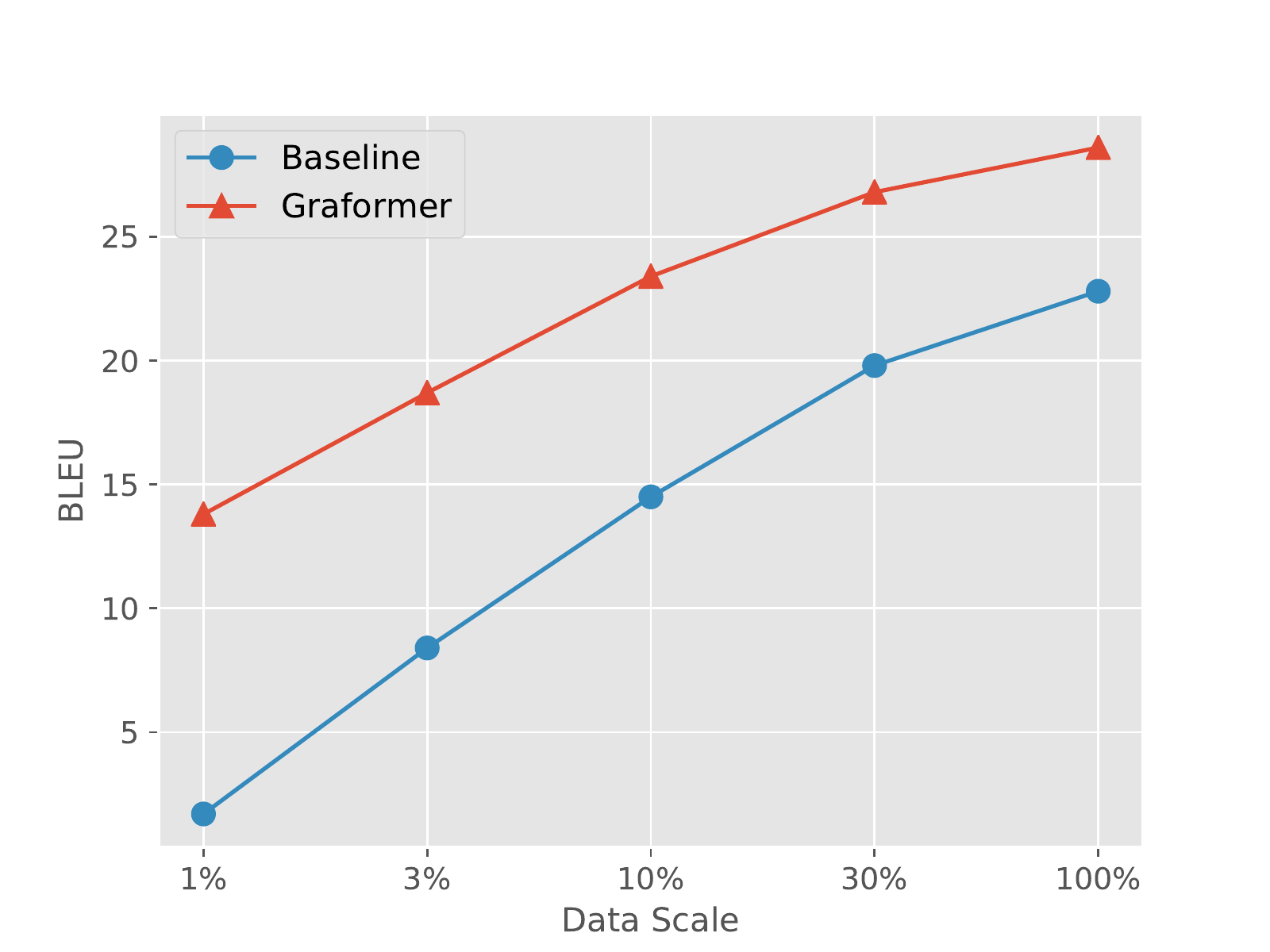}
    \caption{The results of x$\rightarrow$en directions. As the data scale decrease from 100\% to 1\%, the gap is getting larger (5.8$\rightarrow$12.1).}
    \label{fig:x2en}
\end{figure}

\begin{figure}[ht]
    \centering
    \includegraphics[width=0.495\textwidth]{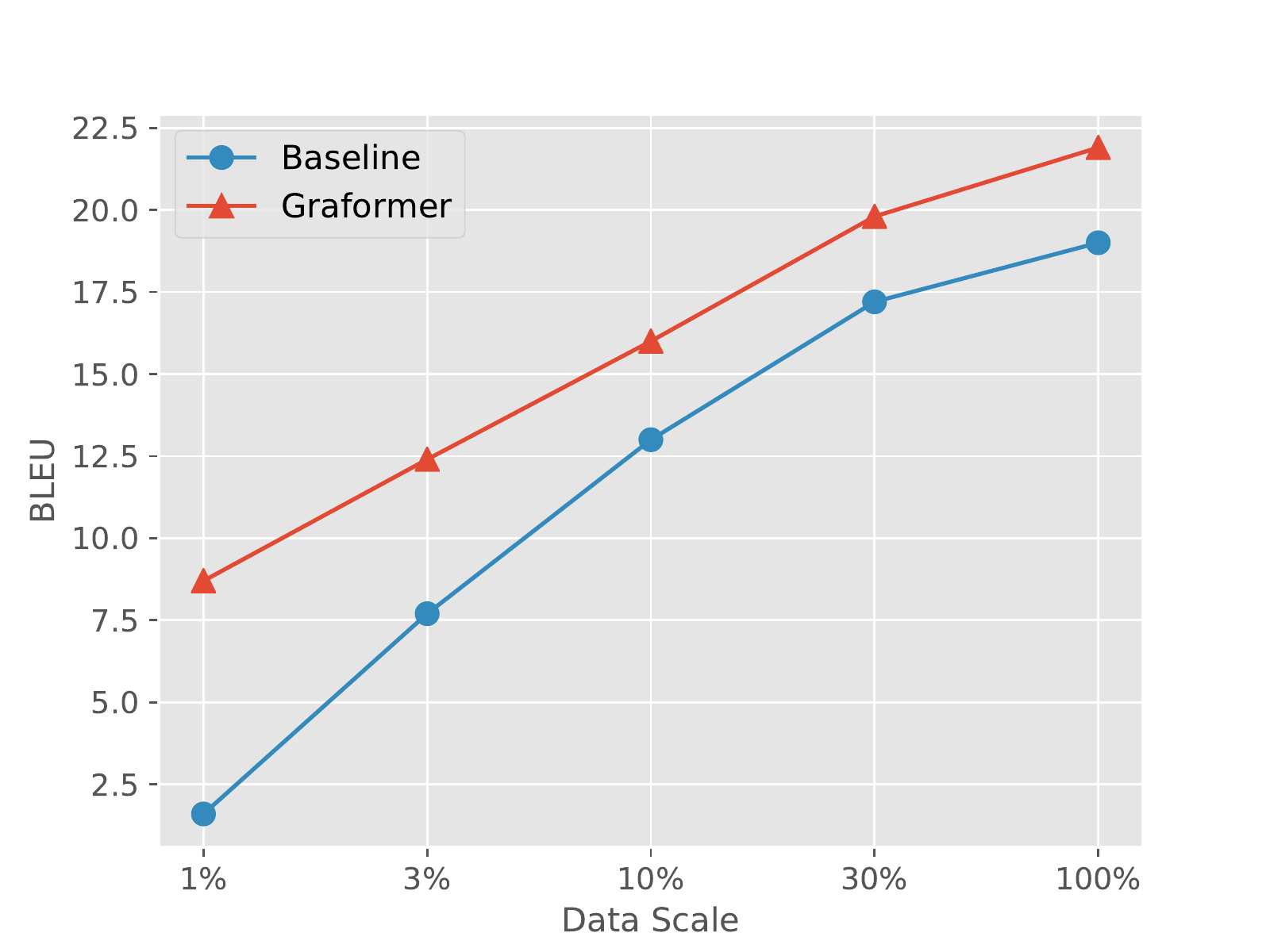}
    \caption{The results of en$\rightarrow$x directions. As the data scale decrease from 100\% to 1\%, the gap is getting larger (2.9$\rightarrow$7.1).}
    \label{fig:en2x}
\end{figure}

We also conduct few-shot experiments. We randomly select 30\%, 10\%, 3\%, 1\% of the data and reproduce the experiments. As is in Figure~\ref{fig:x2en},\ref{fig:en2x}, as the scale of datasets decreases, the performance of baseline drops dramatically and fails to generate comprehensible sentences (BLEU < 5). However, our method keeps relatively higher results even with only 1\% data. And with the less data provided, the gap between \model and baseline is much larger (5.8$\rightarrow$12.1, 2.9$\rightarrow$7.1). Again, it proves that the usage of multiple monolingual data can benefit MNMT greatly since its scale is tens or hundreds of times of the parallel one.

\subsection{The More Monolingual data, The Better}

\begin{figure}[h]
    \centering
    \includegraphics[width=0.495\textwidth]{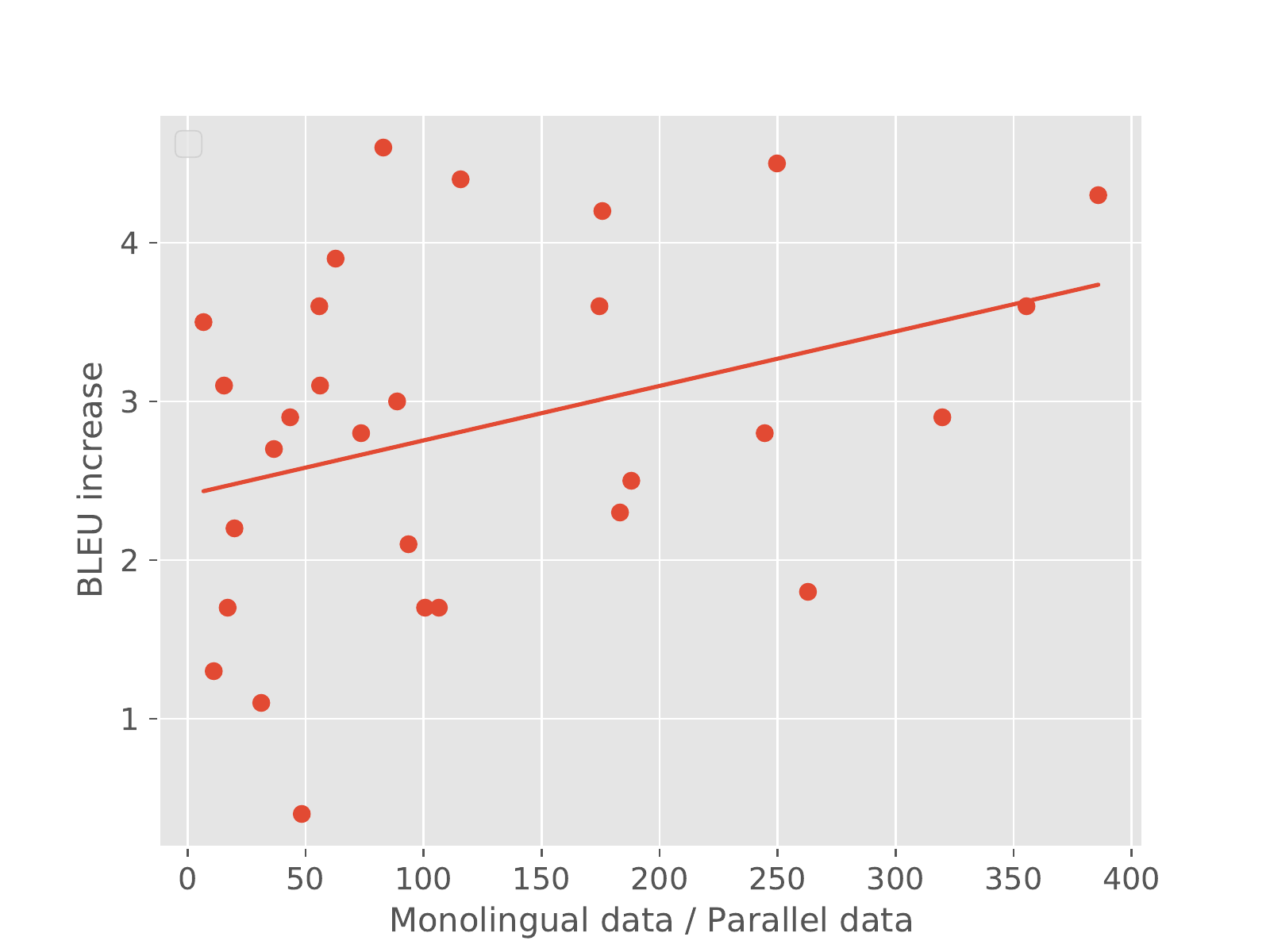}
    \caption{Each point represents a language. The x-axis means the quotient of the monolingual data scale divided by the parallel data scale. The y-axis means the BLEU improvements of en$\rightarrow$x directions.}
    \label{fig:bleu}
\end{figure}

To further analyze the effectiveness of monolingual data, we look into the relationship between the BLEU advance and the data scale. As is in Figure~\ref{fig:bleu}, as the quotient of the monolingual data scale divided by the parallel data scale increases, the BLEU improvements gradually go up. It shows the extra benefit provided by the monolingual data, especially in the large-scale scene. Since the parallel data is rare, \model can be an essential approach to enhance low-resource language pairs.

\begin{table}[h]
    \centering
    \begin{tabular}{c|l|cc}
        \Xhline{3\arrayrulewidth}
        \multirow{2}{*}{Train} & 
        \multirow{2}{*}{Model} & \multicolumn{2}{c}{Testing} \\ 
        \cline{3-4} 
        &  & de$\rightarrow$en & fr$\rightarrow$en \\ 
        \hline
        \multirow{3}{*}{de$\rightarrow$en} & Transformer & 31.9 & 6.7 \\ 
         & \model & \textbf{33.4} & 15.2 \\ 
         & $\text{\model}_{fe}$ & 33.0 & \textbf{20.1} \\ 
        \hline
        \multirow{3}{*}{fr$\rightarrow$en} & Transformer & 5.1 & 35.1 \\ 
         & \model & 10.8 & \textbf{36.0} \\
         & $\text{\model}_{fe}$ & \textbf{16.8} & 35.5  \\
         \Xhline{3\arrayrulewidth}
    \end{tabular}
    \captionsetup{font={footnotesize}}
    \caption{Zero-shot experiments on WMT Datasets. ``$fe$'' means freezing the pre-trained encoder. Notice that even the model does not see parallel sentences for a testing language, our method can achieve 11.7 and 13.4 BLEU improvement.}
    \label{tab:zero_wmt}
\end{table}

\begin{table}[!h]
    \centering
    \begin{tabular}{c|l|cc}
        \Xhline{3\arrayrulewidth}
        \multirow{2}{*}{Train} & 
        \multirow{2}{*}{Model} & \multicolumn{2}{c}{Testing} \\ 
        \cline{3-4} 
        &  & de$\rightarrow$en & fr$\rightarrow$en \\ 
        \hline
        \multirow{3}{*}{de$\rightarrow$en} & Transformer & 33.6 & 1.7 \\ 
         & \model & \textbf{36.9} & 3.4 \\ 
         & $\text{\model}_{fe}$ & 35.4 & \textbf{11.9} \\ 
        \hline
        \multirow{3}{*}{fr$\rightarrow$en} & Transformer & 1.5 & 37.3 \\ 
         & \model & 4.5 & \textbf{40.7} \\
         & $\text{\model}_{fe}$ & \textbf{10.7} & 39.8 \\
         \Xhline{3\arrayrulewidth}
    \end{tabular}
    \captionsetup{font={footnotesize}}
    \caption{Zero-shot experiments on TED Datasets. $``fe''$ means freezing the pre-trained encoder. Notice that even the model does not see parallel sentences for a testing language,  our method can achieve 10.2 and 9.2 BLEU improvement.}
    \label{tab:zero_ted}
\end{table}

\subsection{\model Boosts Zero-Shot Translation}

To verify whether the multilingual pre-trained model learns cross-lingual knowledge, we also conduct a crossed experiment of zero-shot translation. Firstly, we use our approach to train models only on German-English corpus and then conduct inference on French-English test sets. Converse ones are done similarly. We perform experiments on both TED and WMT datasets, with the encoder frozen ($\text{\model}_{fe}$) and tuned (\model).

As is in Table~\ref{tab:zero_wmt} and \ref{tab:zero_ted}, we can draw similar conclusions. On the one hand, the performance of the original direction is improved, as expected. On the other hand, the inference results in the other direction are also significantly boosted at the same time. It is worth noting that our models are trained with none of the test directions but obtain BLEU score high than 10.

More specifically, if the encoder is frozen, the results of the main direction can be slightly lowered, but the results of the zero-shot one will be significantly improved. It illustrates that the untuned pre-trained model contains much more cross-lingual knowledge and can be better transferred to untrained pairs.

\subsection{\model Works in Bilingual Translation}

To verify the effect of our methods, we also conduct experiments on bilingual translation. We use WMT14  English-German and English-French Datasets. In this series of settings, the datasets and vocabulary of both pre-training and tuning are limited in the bilingual corpus. For en-fr training, we adopt $dropout=0.1$, following~\newcite{vaswani2017attention}. 

The results, along with several strong related pre-training works, are listed in Table~\ref{tab:bilingual}. Those related works all take advantage of pre-trained models and significantly improve the translation. Our method boosts the performance of bilingual translation and is at the top level. It proves the universal effectiveness of \model.

\begin{table}[ht]
    \centering
    \begin{tabular}{l|cc}
        \Xhline{3\arrayrulewidth}
        Model & en$\rightarrow$de & en$\rightarrow$fr \\
        \hline
        Transformer & 28.9 & 41.8 \\
        \newcite{yang2020towards} & 30.1 & 42.3 \\
        \newcite{weng2020acquiring} & 29.2 & - \\
        \newcite{yang2020alternating} & 29.2 & - \\
        \newcite{zhu2020incorporating} & 30.8 & \textbf{43.8} \\
        \newcite{rothe2020leveraging} & 30.6 & - \\
        \newcite{guo2020incorporating} & 30.6 & 43.6 \\
        \hline
        \model & \textbf{31.0} & 43.6 \\
        \Xhline{3\arrayrulewidth}
    \end{tabular}
    \caption{Bilingual translation results of English-German and English-French of WMT14. Comparing objects are strong results reported by recent works. \model boosts the performance and is at the top level.}
    \label{tab:bilingual}
\end{table}



\section{Conclusion}
\label{sec:conclusion}

In this paper, we propose \model, grafting multilingual BERT and multilingual GPT for multilingual neural machine translation. By pre-training the representation part (encoder) and generation part (decoder) of the model, we leverage the monolingual data to boost the translation task. And different from other previous fusing methods, we maintain the original architectures. With this approach, we can fully take advantage of the pre-trained models, including their well-trained capacity for representation and generation. Experimental results show that our method can significantly improve the performance and outperform similar related works. A series of empirical analyses of perplexity, few-shot translation, and zero-shot translation also shows its universality.

\bibliography{emnlp2021}
\bibliographystyle{acl_natbib}

\clearpage
\appendix

\section{Languages and Scales}
\label{apdx:langs}

The languages of datasets are listed in Table~\ref{tab:langs} and Table~\ref{tab:pairs}, for pre-training and translation training, respectively. We use significantly less data than mBART\cite{liu2020multilingual}. According to its paper (and some naive summation), they use 208 billion tokens in 1.4T in total. We only use 42 billion tokens in 0.18T.

\section{Results of Fused Style Methods}
\label{apdx:assembled}

The results of other Fused Style methods are in Table~\ref{tab:x2en_appendix} and \ref{tab:en2x_appendix}.

\begin{table}[ht] 
    \begin{tabular}{lr|lr}
        \Xhline{3\arrayrulewidth}
        Language & Scale & Language & Scale \\
        \hline
        am & 119643 & ky & 279440 \\
        bg & 38305118 & lt & 4992036 \\
        bn & 3916068 & lv & 13059185 \\
        bs & 1955342 & mk & 209389 \\
        cs & 90149511 & ml & 182467 \\
        de & 329456604 & mr & 325364 \\
        el & 8159512 & nl & 1205639 \\
        en & 326422361 & or & 444212 \\
        es & 65422557 & pa & 218067 \\
        et & 7023190 & pl & 14480947 \\
        fa & 1304611 & ps & 948310 \\
        fi & 23127824 & pt & 9260529 \\
        fr & 121133895 & ro & 21285406 \\
        gu & 535156 & ro* & 20509504 \\
        hi & 32491838 & ru & 94788355 \\
        hr & 6718607 & so & 168710 \\
        hu & 40181635 & sr & 3798788 \\
        it & 39682711 & sw & 455488 \\
        iu & 781877 & ta & 1251716 \\
        ja & 19579066 & te & 882347 \\
        kk & 1956205 & tr & 17494020 \\
        km & 4410059 & uk & 1486906 \\
        kn & 502499 & zh & 25401930 \\
        \hline
         &  & all & 1.40B \\
        \Xhline{3\arrayrulewidth}
    \end{tabular}
    \caption{Languages used for pre-training and their scales (in sentences). ``ro*'' means processed Romanian.}
    \label{tab:langs}
\end{table}

\begin{table}[ht] 
    \begin{tabular}{lr|lr}
        \Xhline{3\arrayrulewidth}
        Language & Scale & Language & Scale \\
        \hline
        bg & 174444 & ja & 204090 \\
        bn & 4649 & kk & 3317 \\
        bs & 5664 & lt & 41919 \\
        cs & 103093 & mk & 25335 \\
        de & 167888 & mr & 9840 \\
        el & 134327 & nl & 183767 \\
        es & 196026 & pl & 176169 \\
        et & 10738 & pt & 51785 \\
        fa & 150965 & ro & 180484 \\
        fi & 24222 & ru & 208458 \\
        fr & 192304 & sr & 136898 \\
        hi & 18798 & ta & 6224 \\
        hr & 122091 & tr & 182470 \\
        hu & 147219 & uk & 108495 \\
        it & 204503 & zh & 5534 \\
        \hline
         & & all & 3.18M \\
        \Xhline{3\arrayrulewidth}
    \end{tabular}
    \caption{Language pairs (from \& to English) used for translation training and their scales (in sentences).}
    \label{tab:pairs}
\end{table}

\begin{table*}[ht]
    \centering
    \begin{tabular}{l|cccccccccc}
        \Xhline{3\arrayrulewidth}
        Model & bg & bn & bs & cs & de & el & es & et & fa & fi \\ \hline
        Direct & 36.8 & 18.0 & 35.2 & 28.5 & 33.9 & 35.3 & 39.1 & 22.5 & 24.8 & 21.2 \\
        Adapter & 38.0 & 18.1 & 36.8 & 29.2 & 34.3 & 36.2 & 40.1 & 23.3 & 23.7 & 21.9 \\
        \model & \textbf{38.5} & \textbf{18.1} & \textbf{36.5} & \textbf{29.4} & \textbf{35.5} & \textbf{37.4} & \textbf{40.7} & \textbf{24.0} & \textbf{26.9} & \textbf{23.0} \\
        \Xhline{3\arrayrulewidth}
        Model & fr & hi & hr & hu & it & ja & kk & lt & mk & mr \\ \hline
        Direct & 38.0 & 23.6 & 35.3 & 24.4 & 36.0 & 12.3 & 10.1 & 25.4 & 33.8 & 12.1 \\
        Adapter & 38.7 & 24.1 & 36.2 & 24.9 & 36.5 & 11.7 & 10.1 & 25.9 & 34.7 & 11.2 \\
        \model & \textbf{39.2} & \textbf{25.1} & \textbf{36.7} & \textbf{26.1} & \textbf{37.2} & \textbf{13.7} & \textbf{10.5} & \textbf{27.2} & \textbf{35.7} & \textbf{13.0} \\
        \Xhline{3\arrayrulewidth}
        Model & nl & pl & pt & ro & ru & sr & ta & tr & uk & zh \\ \hline
        Direct & 33.2 & 23.6 & 40.1 & 33.6 & 23.9 & 33.9 & 8.7 & 23.3 & 27.8 & 18.5 \\
        Adapter & 33.8 & 24.0 & 41.1 & 34.2 & 24.3 & 34.9 & 7.1 & 22.9 & 27.6 & 17.9 \\
        \model & \textbf{35.2} & \textbf{25.1} & \textbf{41.5} & \textbf{35.1} & \textbf{25.1} & \textbf{35.6} & \textbf{10.2} & \textbf{25.5} & \textbf{28.9} & \textbf{19.9} \\
        \Xhline{3\arrayrulewidth}
    \end{tabular}
    \caption{The results of x$\rightarrow$en directions for ``Direct''~\cite{rothe2020leveraging,ma2020xlm} and ``Adapter''~\cite{guo2020incorporating}.}
    \label{tab:x2en_appendix}
\end{table*}

\begin{table*}[ht]
    \centering
    \begin{tabular}{l|cccccccccc}
        \Xhline{3\arrayrulewidth}
        Model & bg & bn & bs & cs & de & el & es & et & fa & fi \\ \hline
        Direct & 30.7 & 12.2 & 24.5 & 18.2 & 25.1 & 27.8 & 35.3 & 14.9 & 13.3 & 13.1 \\
        Adapater & 31.0 & 10.5 & 24.3 & 18.5 & 25.4 & 26.9 & 35.3 & 15.3 & 9.6 & 13.4 \\
        \model & \textbf{33.0} & \textbf{14.1} & \textbf{26.3} & \textbf{20.2} & \textbf{27.8} & \textbf{29.8} & \textbf{37.5} & \textbf{16.1} & \textbf{14.2} & \textbf{14.4} \\
        \Xhline{3\arrayrulewidth}
        Model & fr & hi & hr & hu & it & ja & kk & lt & mk & mr \\ \hline
        Direct & 35.4 & 16.7 & 25.2 & 15.7 & 30.7 & 12.2 & 4.0 & 14.5 & 24.4 & 10.5 \\
        Adapater & 35.8 & 15.3 & 24.8 & 15.7 & 30.2 & 9.2 & 3.8 & 14.5 & 24.5 & 9.0 \\
        \model & \textbf{37.8} & \textbf{18.1} & \textbf{26.8} & \textbf{17.2} & \textbf{32.5} & \textbf{12.8} & \textbf{3.8} & \textbf{15.9} & \textbf{25.7} & \textbf{10.6} \\
        \Xhline{3\arrayrulewidth}
        Model & nl & pl & pt & ro & ru & sr & ta & tr & uk & zh \\ \hline
        Direct & 28.1 & 14.4 & 34.3 & 26.9 & 17.5 & 20.2 & 15.6 & 12.3 & 18.7 & 21.8 \\
        Adapater & 26.9 & 13.9 & 34.2 & 26.7 & 17.1 & 19.7 & 11.6 & 11.7 & 18.1 & 20.5 \\
        \model & \textbf{29.0} & \textbf{15.8} & \textbf{36.6} & \textbf{29.1} & \textbf{19.0} & \textbf{21.4} & \textbf{14.7} & 13.3 & \textbf{19.5} & \textbf{23.0} \\
        \Xhline{3\arrayrulewidth}
    \end{tabular}
    \caption{The results of en$\rightarrow$x directions for ``Direct''~\cite{rothe2020leveraging,ma2020xlm} and ``Adapter''~\cite{guo2020incorporating}.}
    \label{tab:en2x_appendix}
\end{table*}

\end{document}